# Automatic Road Lighting System (ARLS) Model Based on Image Processing of Moving Object

[1]Suprijadi, [2]Thomas Muliawan, [3]Sparisoma Viridi

[1] Assoc. Prof., Department of Physics, Institut Teknologi Bandung, Bandung 40132, Indonesia

[2] Undergraduate Student, Department of Physics, Institut Teknologi Bandung, Bandung 40132, Indonesia

[3] Asstt. Prof., Department of Physics, Institut Teknologi Bandung, Bandung 40132, Indonesia

E-mail: [1]supri@fi.itb.ac.id, [3]dudung@fi.itb.ac.id

**ABSTRACT**

Using a vehicle toy (in next future called vehicle) as a moving object an automatic road lighting system (ARLS) model is constructed. A digital video camera with 25 fps is used to capture the vehicle motion as it moves in the test segment of the road. Captured images are then processed to calculate vehicle speed. This information of the speed together with position of vehicle is then used to control the lighting system along the path that passes by the vehicle. Length of the road test segment is 1 m, the video camera is positioned about 1.1 m above the test segment, and the vehicle toy dimension is 13 cm × 9.3 cm. In this model, the maximum speed that ARLS can handle is about 1.32 m/s, and the highest performance is obtained about 91 % at speed 0.93 m/s.

## 1. INTRODUCTION

Road lighting system, especially in the night or in bad weather, plays important role in safety of people that is using the street: driving and walking [1]. Unfortunately, in a big city like Jakarta, Indonesia, there are 205.852 lamp units registered which operate about 12 hours/day and cost 142.3 billion IDR for their operation in 2007, and in 2010 additional 262 thousand points will be executed [2]. This cost can be reduced to 76 % using Dimmable Road Lighting System, e.g. between 06.00 PM – 06.00 AM, where only some of road lamps are operated depending on load of the traffic [3], but it still dangerous and not responsible since we can not predict whether there are vehicles or not at that time passing the road. Other solution is to build an automatic road lighting system (ARLS) [4] which can detect existence of vehicle on the road and then switch the lamps on or of according in where direction the vehicle moves. Many light control systems were developed, for example based on infra red [5], time schedule [6], or other sensors [7].

In the last decade, image processing techniques were growing fast. The supporting by a new microprocessor and hardware, the technologies were applied in many filed including for control and security. Some applications of image processing were reported by authors in Indonesian journal or national conferences [8-9].

The other opportunity in control system sensor can be replaced by an image from digital camera [10-12]. We proposed the opportunity to replace an usual sensor with an image to make a control system. We developed an ARLS model using image processing as main control system. The model consists of vehicle toy, video camera as input, processing unit to calculate and predict vehicle motion parameters, and controller to switch the lamps on and off. A vehicle toy is used as the motion object.

## 2. SYSTEM SETUP

Automatic road lighting system (ARLS) consists of four main parts: vehicle toy as moving object, video camera as input part, a personal computer (PC) as processing part, and electronic actuator circuit 555 based as output part with road lamp. Relations between these parts are shown in Figure 1.

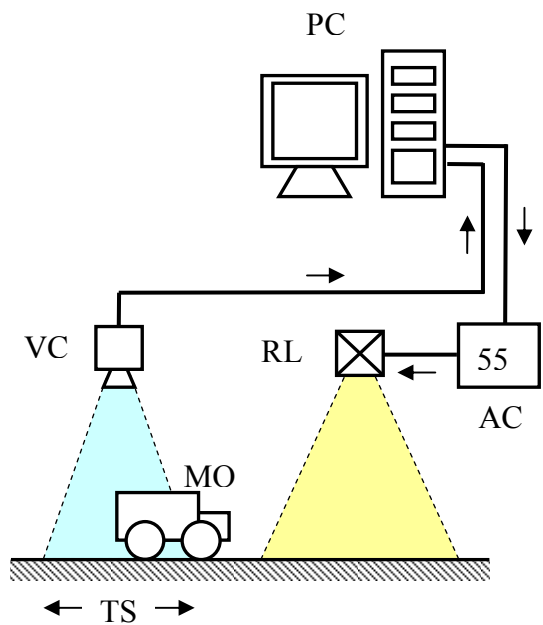

**Figure 1**: Illustration of ARLS: vehicle as moving object (MO), video camera (VC) which operates at road test segment (TS) as input part, a personal computer (PC) as processing part, and electronic actuator circuit 555 based (AC) which controls road lamp (RL) as output

A typical webcam with 25 fps specification is used as input part of the ARLS. It feeds the processing part (PC) with series of captured images as the vehicle toy (MO) is passing the road test segment (TS). As the processing unit a PC with Intel® CoreTM 2 6400 processor, 2.13 GHz clock speed, FSB 133.2 MHz, L2 cache 3.2 kB, and 1014 MB DDR-SDRAM memory. For the operating system a Microsoft Windows XP Professional (2002) with Service Pack 2 is used.

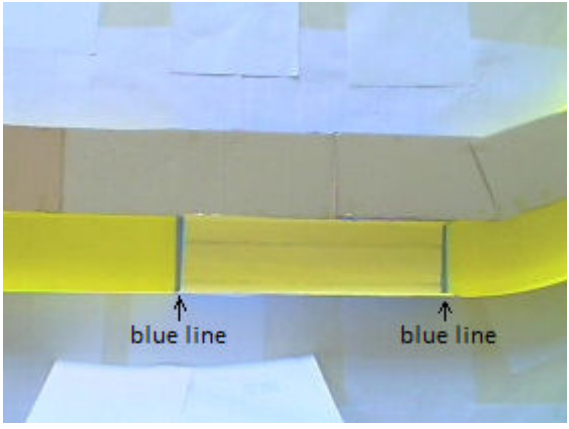

**Figure 2**: Original captured image of road test segment (TS) without passing vehicle toy which is used as default background image in processing the series of capture images as the vehicle is passing this TS

In order to calibrate distance in captured image two blue lines are drawn as illustrated in Figure 2. One of the image processing tasks is to distinguish these two lines out of other object in captured image. If distance in real world is $l$ and in the captured image is $L$, then a scaling factor $c$ can be defined

$$c = \frac{l}{L}, \quad (1)$$

which is used to transform all positions of pixels in the captured image $(X, Y)$ which are represented in pixels to $(x, y)$ which are represented in m through

$$\begin{pmatrix} x \\ y \end{pmatrix} = c \begin{pmatrix} X \\ Y \end{pmatrix}, \quad (2)$$

Value of $c$ is dependent on position of VC above the TS. The next task of image processing is to identify the MO as it is passing the TS. The identifying process is performed by subtracting pixels value a captured image at time $t$ with a referenced image (indeks $R$) which the background image (Figure 2) is, but the subtracting is forced to be 1 if value of the pixels in a captured image is different than value of pixels in referenced image. This information is place in a new image (indeks $N$),

$$\begin{pmatrix} X_N \\ Y_N \end{pmatrix}^{(t)} = \text{ceila} \left[ \begin{pmatrix} X \\ Y \end{pmatrix}^{(t)} - \begin{pmatrix} X_R \\ Y_R \end{pmatrix} \right], \quad (3)$$

where the ceila($x$) function (or ceil from absolute) is defined as

$$\text{ceila}(x) = \begin{cases} 0, & x = 0 \\ 1, & x \neq 0 \end{cases}. \quad (4)$$

The next step is to find the center of MO in every captured image. It can be done by putting incremental value $I_x$ from 1 to some value for $X$ as the new image scanned in horizontal direction and $I_Y$ for $Y$ in vertical direction. Illustration for rectangular object is given Figure 3. Then center of MO is found using

$$X_{MO} = \frac{\max(I_X) - \min(I_X)}{A}, \quad (5)$$

$$Y_{MO} = \frac{\max(I_Y) - \min(I_Y)}{A}, \quad (6)$$

where are area is found as number of steps needed to cover all pixels in the new image that have value 1. This algorithm holds also for non-rectangular form.

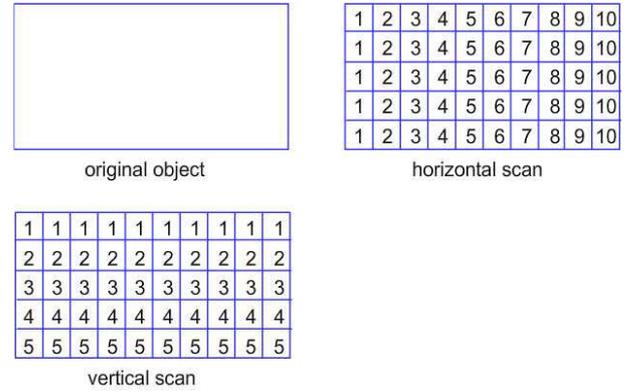

**Figure 3**: A sample of rectangular form in new image for $I_X = 1..10$ and $I_Y = 1..5$ with $A = 50$ (all are in pixels unit)

The displacement of MO for time difference is then found to be

$$\Delta r = c \sqrt{\left[ X_{MO}^{(t+\Delta t)} - X_{MO}^{(t)} \right]^2 + \left[ Y_{MO}^{(t+\Delta t)} - Y_{MO}^{(t)} \right]^2}, \quad (7)$$

by using the scaling factor in Equation (1), that makes $\Delta r$ already in m. Equation (5) tells us that the displacement holds also for non-vertical or non-horizontal motion, but in $\Delta t$ the MO is considered has a straight motion. Speed of the MO is then obtained by dividing Equation (7) with $\Delta t$, which is

$$v = \frac{\Delta r}{\Delta t}, \quad (8)$$

The last step of processing part is to calculate or predicted position of MO in the future, for example at $t + \Delta t'$ position of MO is at

$$r(t + \Delta t') = r(t) + v \Delta t', \quad (9)$$

where $r$ is distance along the path of the road. Using this information the road lamps that are positioned at $t + \Delta t'$, $t + \Delta t''$, .. can be switched on and off as they are needed to. The processing unit will send a signal to AC to turn the light on or of at some distance $r$ the position of the VC.

The algorithm:

```
Start
N ← took image
N +1 ← took 2nd image
'detect MO, 1st
   while (N + 1) - N ≠ 0
      switch on (trigger)
'detect MO, 2nd in 2nd line
   while (N + 1) - N ≠ 0
      switch off (reset)
```

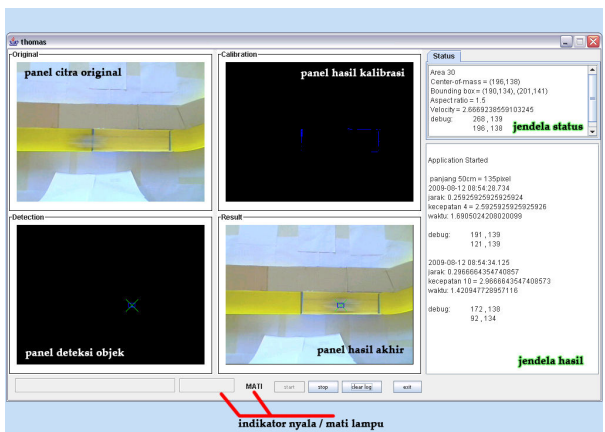

**Figure 4**: User interface of program used in ARLS consists of live captured image window, scaling factor (calibration) window, MO position detection, overlay (original image + MO position) window, motion parameter window, and calculation result and actuating window

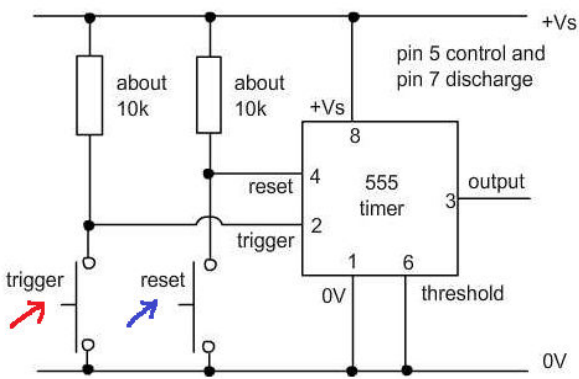

**Figure 5**: Electronic actuator circuit 555 based (AC) is a bistable multivibrator whose states are controlled by a signal fed to trigger and reset pins

In order to process the input, to calculate the displacement and speed of MO, and to actuate the road lamps a program written in Java is built, whose user interface is shown in Figure 4.

The AC which is based on a bistable multivibrator using NE555 is illustrated in Figure 5. This electronic actuator job is to receive signal from processing part and that switch the road lamp on or off. Since ARLS designed here using a parallel port [13] to actuate the road lamps, it is possible to control up to 8 lamps. Further detail of 555 based electronic circuit can be found in some sources [14-15].

## 3. RESULTS AND DISCUSSION

The ARLS is illustrated in Figure 6. The calibration step with position of VC about 1.075 m above the TS and length of TS about 1 m gives $c$ = 2.7 cm/pixels.

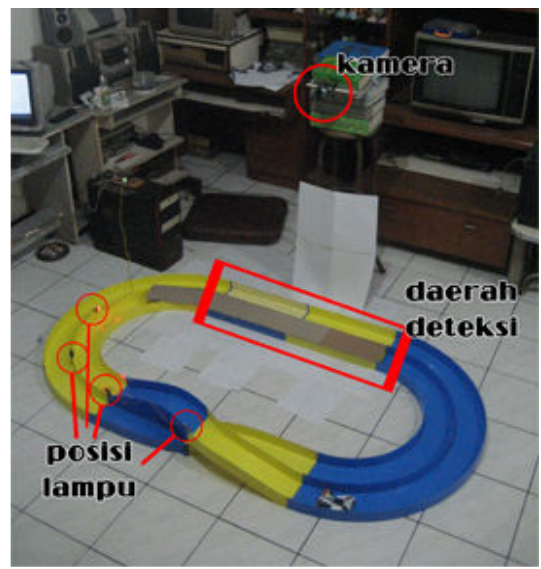

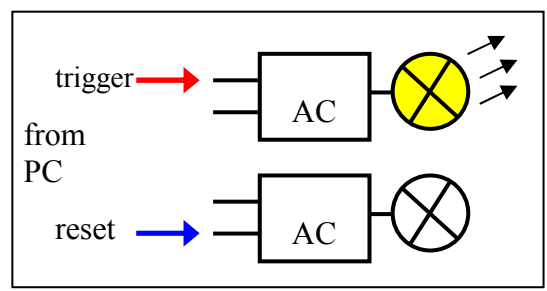

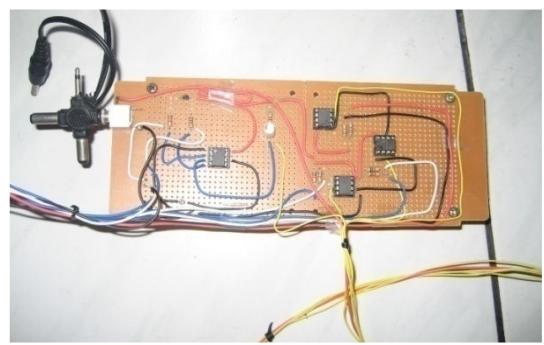

**Figure 6**: Whole system of ARLS: vehicle toy tracks, VC, PC, and TS (top), the AC part with cables for road lamps (RL) (middle), and function of trigger and reset pins on AC (bottom)

Table 1: Measurements variable using stopwatch (index S) and ARLS, image quality, and performance of ARLS

| $\langle v_S \rangle$ (m/s) | $\langle v_{ARLS} \rangle$ (m/s) | Image quality (% blur) | $|\Delta v|$ (%) | $r_{MO} - r_{RL}$ | Performance (%) |
|---|---|---|---|---|---|
| 0.93 | 1.03 | 10 | 10.75 | ~ 0 | 91 |
| 1.32 | 1.63 | 25 | 23.48 | ~ 0 | 74 |
| 1.41 | 1.85 | 33 | 31.21 | < 0 | 54 |
| 1.52 | 2.19 | 40 | 44.08 | < 0 | 36 |
| 2.03 | 1.13 | > 50 | 44.33 | > 0 | 0 |

To check the result of speed calculation for ARLS ($\langle v_{ARLS} \rangle$) an observation using bare eye and stopwatch is conducted and results are labeled $\langle v_S \rangle$ as shown in Table 1. Both speeds are averaged from several observations. Since the speed of MO is varied, some captured image (image quality) is so blur that can not be further processing or gives a bad results. The performance of ARLS is counted by doing a 100 trials and the number of success is registered. For example, 91 % performance mean that 91 success out of 100 trials. The parameter $r_{MO} - r_{RL}$ is used to indicated whether ARLS switch on the light as the MO pass the lamp (~ 0), or before (< 0), or after (> 0).

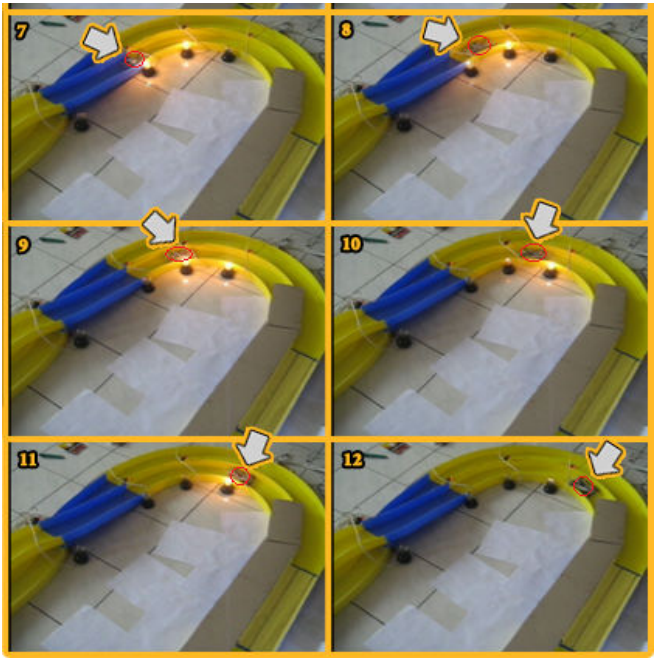

Figure 7: Illustration of good performance of switched on RL (white spots) as MO (indicating by white arrow) is passing it.

It can bee seen from Figure 8 that correlation between vS and vARLS in range of ±45 %. The jump at about 2 m/s velocity is due the low PC execution time compared to the velocity of the vehicle. Further investigation is needed to confirm the performance dependence MO speed which should not arise in a ARLS.

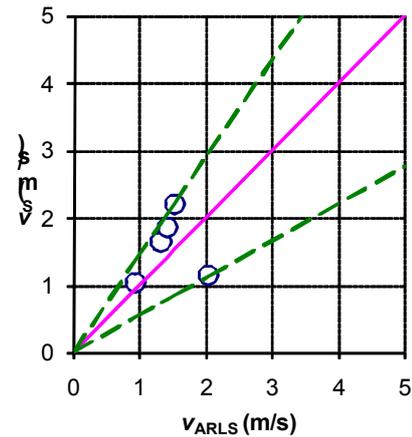

Figure 8: Correlation of measured speed of the vehicle toy (MO) using stopwatch (index S) and ARLS (index ARLS) in range of ±45 % (dashed lines) of equality (solid line)

## 4. CONCLUSION

The highest performance of our automatic road lighting system (ARLS) model, which is about 91 %, is measured at vehicle toy speed 0.93 m/s with camera height 1.075 m, video recording setting 25 fps, road length of test segment 1 m, and vehicle toy dimension 13 cm × 9.3 cm. By setting the limit of acceptable good performance at value above 70%, with previous configuration, the maximum speed of vehicle toy allowed in the ARLS model is about 1.32 m/s. Distance calibration in several road segments are urgently needed and must be not only relied on color distinction since it is really dependent on ambient light which is different in every road segment and can be controlled and predicted. There is no correlation between vS and vARLS observed but they are out of proportional in range of ±45 %.

## AUTHOR PROFILES

**1. Suprijadi** received the bachelor degree in physics from Institut Teknologi Bandung, Indonesia, in 1991. He received the master degree in Quantum Engineering from the Nagoya University - Japan, in 1998. The doctor degree in Engineering is awarded to him in 2001 from Nagoya University, Japan. Currently, he is an Associate Professor at Physics Department, Institut Teknologi Bandung, Indonesia and working in field of instrumentation, computation, and nano material design.

**2. Thomas Muliawan** was an undergraduate student of Physics Department, Institut Teknologi Bandung.

**3. Sparisoma Viridi** received the bachelor degree in physics from Institut Teknologi Bandung, Indonesia, in 1998. He received the master degree in the same field from the same university, in 2001. The doctor degree in natural science is awarded to him in 2007 from Dortmund University, Germany. Currently, he is an Assistant Professor at Physics Department, Institut Teknologi Bandung, Indonesia and working in field of granular physics, fluids, and computation.